\documentclass[letterpaper, 10 pt, conference]{ieeeconf}  

\IEEEoverridecommandlockouts                              

\usepackage{epsfig}
\usepackage{makecell}
\usepackage{subcaption}
\usepackage{xcolor}
\usepackage{ragged2e}
\usepackage{float} 

\usepackage{graphicx}
\usepackage{amsmath}
\usepackage{adjustbox}
\usepackage{amssymb}
\usepackage{booktabs}
\usepackage{tabularx}
\usepackage{array}
\usepackage{makecell}
\usepackage{bm}

\usepackage{kantlipsum}
\usepackage{graphicx}
\usepackage{etoolbox}
\newcommand{\insertfig}{%
  \includegraphics[width=\linewidth, height=220pt]{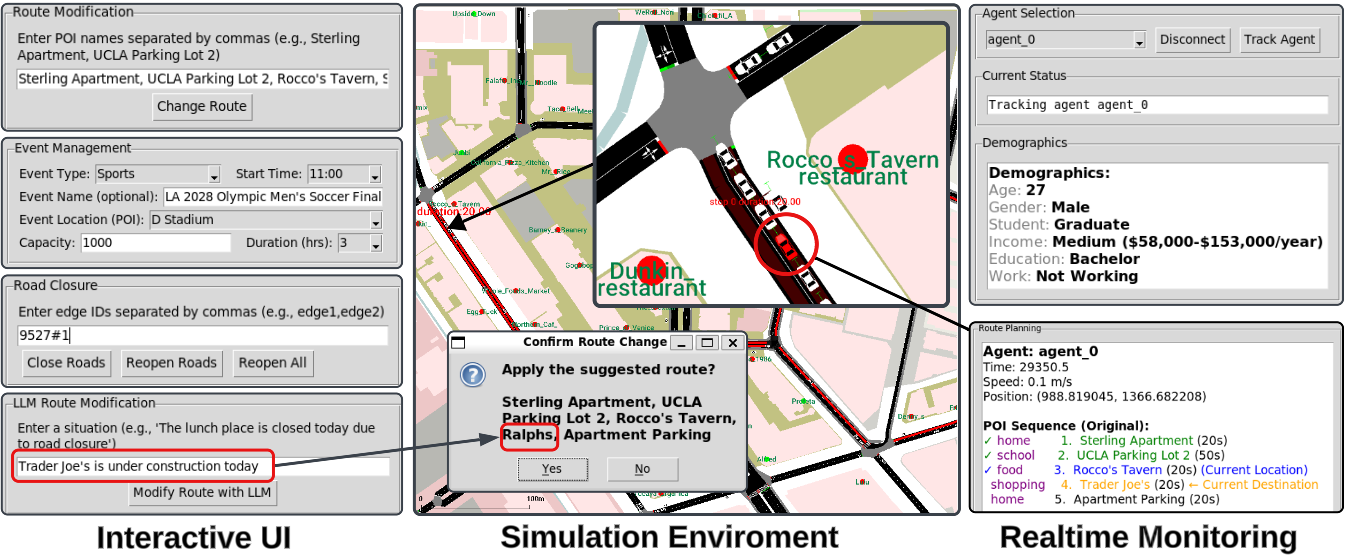}%
  \captionof{figure}{MobiVerse visualization interface: Users can observe agent behaviors in the simulation view, track individual agents, set road closures, introduce gathering events, or directly communicate with agents to influence their travel decisions and observe adaptation in real time.}%
  \label{fig:flow_diagram_1}%
}

\makeatletter
\apptocmd{\@maketitle}{\centering\insertfig}{}{}
\makeatother

\title{\LARGE \bf
MobiVerse: Scaling Urban Mobility Simulation with Hybrid Lightweight Domain-Specific Generator and Large Language Models
}

\author{Yifan Liu, Xishun Liao$^{*}$, Haoxuan Ma, Jonathan Liu, Rohan Jadhav, and Jiaqi Ma
\thanks{The authors are with the UCLA Mobility Lab under the Department of Civil and Environmental Engineering, University of California, Los Angeles, Los Angeles, USA.}
\thanks{*Corresponding author: xishunliao@ucla.edu}
}

\begin{document}

\maketitle
\thispagestyle{empty}
\pagestyle{empty}

\begin{abstract}
Understanding and modeling human mobility patterns is crucial for effective transportation planning and urban development. Despite significant advances in mobility research, there remains a critical gap in simulation platforms that allow for algorithm development, policy implementation, and comprehensive evaluation at scale. Traditional activity-based models require extensive data collection and manual calibration, machine learning approaches struggle with adaptation to dynamic conditions, and treding agent-based Large Language Models (LLMs) implementations face computational constraints with large-scale simulations. To address these challenges, we propose MobiVerse, a hybrid framework leverages the efficiency of lightweight domain-specific generator for generating base activity chains with the adaptability of LLMs for context-aware modifications. A case study was conducted in Westwood, Los Angeles, where we efficiently generated and dynamically adjusted schedules for the whole population of approximately 53,000 agents on a standard PC. Our experiments demonstrate that MobiVerse successfully enables agents to respond to environmental feedback, including road closures, large gathering events like football games, and congestion, through our hybrid framework. Its modular design facilitates testing various mobility algorithms at both transportation system and agent levels. Results show our approach maintains computational efficiency while enhancing behavioral realism. MobiVerse bridges the gap in mobility simulation by providing a customizable platform for mobility systems planning and operations with benchmark algorithms. Code and videos are available at https://github.com/ucla-mobility/MobiVerse.

\end{abstract}

\section{INTRODUCTION AND RELATED WORK}



Human mobility modeling and simulation plays a crucial role in urban planning, transportation system design, and policy evaluation. By accurately modeling and simulating individual movement patterns, researchers and planners can forecast traffic conditions, assess infrastructure changes, and develop effective mobility solutions \cite{gonzalez2008understanding}.

As the backbone to support the simulation process, human mobility modeling approaches have evolved through different paradigms as computational power and data availability have increased. Traditional activity-based models \cite{arentze2009need, nijland2014multi} depend on predefined rules and extensive data collection from household surveys, census data, and land-use information. While these models offer detailed behavioral modeling, their rigidity in dynamic conditions and substantial manual calibration requirements make them unsuitable for large-scale applications. Additionally, their data-intensive nature creates challenges in data-scarce environments \cite{solmaz2019survey, xu2015understanding}.

Recent advances in machine learning have enabled more efficient modeling of mobility patterns based on historical data \cite{feng2020learning, luca2021survey}. Deep learning approaches have further enhanced prediction capabilities, generating realistic activity patterns that reflect observed human behavior \cite{liao2024deep}. However, simulations based on these domain-specific generators struggle with adaptation to unusual scenarios or changing conditions, limiting their practical application in dynamic urban environments. They excel at reproducing patterns from training data but cannot effectively respond to novel situations or real-time changes in the environment.

Complementing these approaches, Large Language Models (LLMs) have demonstrated promising capabilities for simulating adaptive human behavior by leveraging their knowledge of human behavior, spatial relationships, and temporal patterns \cite{liu2024semantic, gong2024mobility, wang2023would}. Recent work has explored using LLMs to create agents capable of planning and reflection, enabling more realistic responses to dynamic conditions \cite{park2023generative, lin2023agentsims, jiawei2024large, li2024more}. Despite these advances, LLM-based simulations face significant limitations in scalability for urban-scale applications involving tens of thousands of agents.

To realize mobility modeling results in practice, various simulation platforms have been developed, each with distinct focuses. Simulation of Urban MObility (SUMO) \cite{lopez2018microscopic} operates at the microscopic level, emphasizing traffic conditions and car-following models for precise vehicle-by-vehicle simulation. MATSim \cite{w2016multi, he2024multi} focuses on agent-level transportation optimization through utility-based iterative processes, handling tens of thousands to millions of agents. Game engine-based simulators like "Cities: Skylines" \cite{pinos2020automatic} offer comprehensive visual environments but remain difficult to integrate into academic research due to their complexity. Despite their capabilities, existing platforms have significant limitations—most cannot effectively integrate with advanced mobility modeling algorithms like machine learning or LLMs, and typically lack mechanisms for dynamic behavior adaptation and true behaviorial intelligence in response to environmental feedback.

Existing mobility modeling methods present clear trade-offs: learning-based domain-specific generation models efficiently generate activity patterns but struggle with adaptation to unforeseen conditions, while pure LLM frameworks, though highly adaptive, cannot scale effectively for large-scale urban simulations. Furthermore, the absence of a unified, scalable simulation platform has hindered progress in mobility simulation research, as current tools are often specialized for specific aspects and lack the capability to seamlessly incorporate state-of-the-art mobility modeling algorithms.

To bridge this gap, as shown in Fig.~\ref{fig:flow_diagram_1}, we propose MobiVerse, an open-source platform that combines efficient population synthesis and activity chain generation with dynamic behavior modification through LLM-based decision making. MobiVerse seamlessly integrates with SUMO \cite{lopez2018microscopic} for microscopic traffic simulation while supporting up to tens of thousands of agents through its scalable architecture. Our hybrid approach leverages the computational efficiency of lightweight domain-specific generator with the contextual adaptability of LLMs, enabling large-scale simulations that can dynamically respond to environmental feedback, a capability lacking in existing platforms. The main contributions of this paper include:

\begin{itemize}
    \item A hybrid architecture that combines the efficiency of lightweight domain-specific generators for producing base activity chains with the adaptability of LLMs for context-aware modifications
    \item A scalable simulation framework capable of handling tens of thousands agents while maintaining computational efficiency and behavioral realism
    \item Dynamic response capabilities that enable agents to modify their behavior in response to congestion, road closures, and rare events through LLM-based contextual reasoning and real-time environmental feedback integration, providing more realistic simulation outcomes
    \item A real-world case study demonstrating the framework's capabilities in the Westwood area of Los Angeles
\end{itemize}

\section{METHODOLOGY AND SYSTEM ARCHITECTURE}

\subsection{System Architecture and Framework Design}


MobiVerse is developed with a modular architecture that enables easy adaptation to different locations and specific features, while also providing an interactive UI interface for user interaction. As shown in Fig. \ref{fig:flow_diagram_2}, the system implements a hybrid approach through three interconnected computational modules. 

The system consists of three primary modules: (1) A lightweight Domain-Specific \textit{Base Activity Chain Initializer} that generates baseline activity patterns from socio-demographic data, (2) An LLM-empowered \textit{Activity Chain Modifier} that adapts activities based on environmental conditions, and (3) A visualized \textit{Simulation Environment} that executes and monitors agent behaviors, with bi-directional connections enabling real-time feedback and adaptation between components.

\renewcommand{\thefigure}{2}
\begin{figure}[h]
    \centering
    \includegraphics[width=0.95\columnwidth]{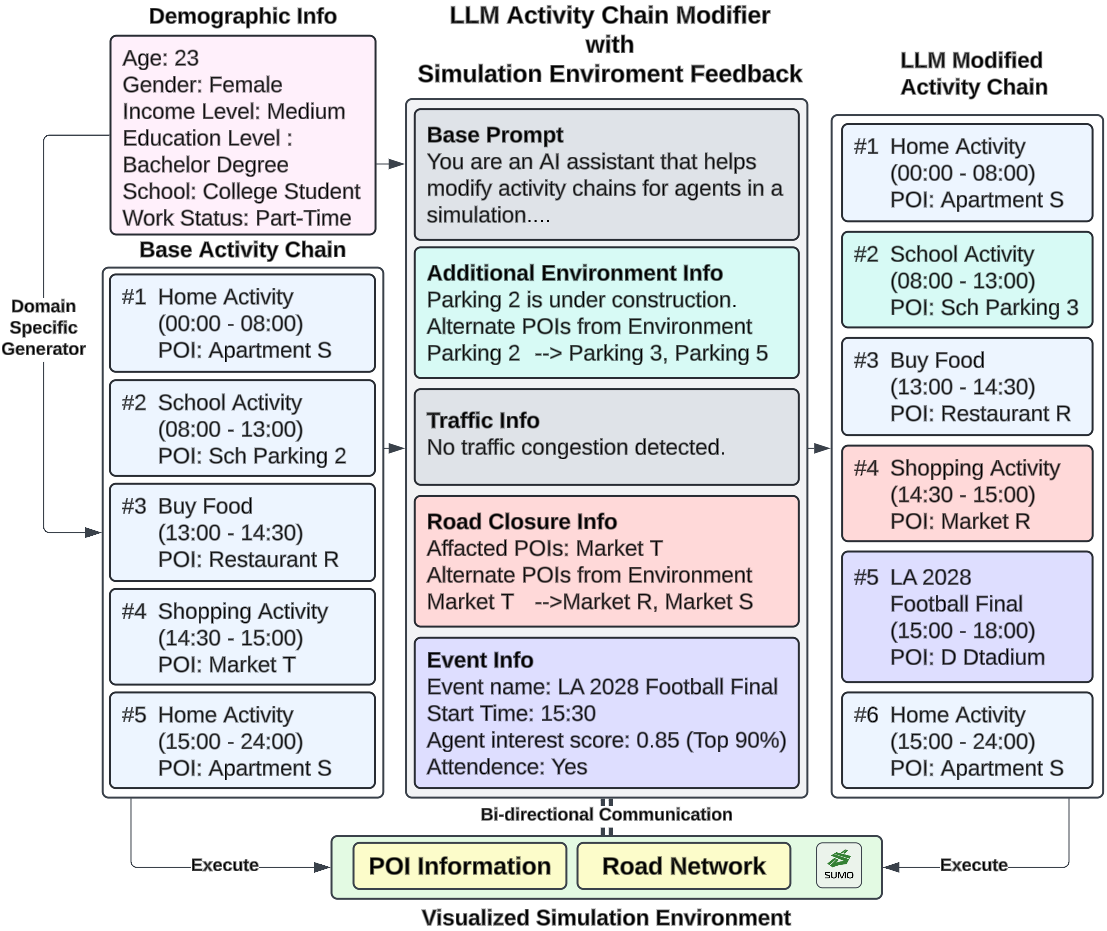}
    \caption{Class diagram of MobiVerse framework showing the main components and their relationships.}
    \label{fig:flow_diagram_2}
\end{figure}
\vspace{-0.3cm}

As shown in Fig.~\ref{fig:flow_diagram_2}, MobiVerse operates through three interconnected modules. The lightweight \textit{Base Activity Chain Initializer} generates baseline activity patterns for all agents using their demographic data as input, producing realistic daily schedules by capturing mobility patterns from historical data and considering factors like age, occupation, and income level. These initialized activity chains are then passed to the \textit{Simulation Environment}, which integrates with SUMO for traffic simulation. When environmental changes occur (road closures, congestion, or special events), the LLM-empowered \textit{Activity Chain Modifier} processes these conditions and adapts agent behaviors accordingly, using structured prompts with environmental, traffic, and event information to generate contextually appropriate modifications to affected agents' activity chains. These modified plans are fed back into the simulation environment through bi-directional communication channels, enabling real-time adaptation. In the following parts of this section, we provide a detailed examination of MobiVerse's core functionalities and technical implementation. 

\subsection{Agent Demographic and Base Activity Chain Initializer}
Each agent in MobiVerse is characterized by a comprehensive socioeconomic profile that includes age, gender, income level, employment status, education level, household composition, and vehicle accessibility. These demographic attributes significantly influence mobility patterns and activity preferences, serving as the foundation for realistic agent behavior simulation. The base \textit{Base Activity Chain Initializer} leverages these demographic characteristics along with available points of interest (POI) information to generate initial daily schedules that reflect typical behavioral patterns.

For each agent $i$ with available demographic information $D_i = \{{d_i}^1, {d_i}^2, \dots, {d_i}^n\}$, we generate a daily activity chain $C_i$ where each activity is defined by its type $A$, start time $T_s$, end time $T_e$, and POI location $P$. The output activity chain $C_i$ for agent $i$ can be represented as 

$C_i = [{A_i}^1, {T_{s,i}}^1, {T_{e,i}}^1, {P_i}^1], \dots, [{A_i}^n, {T_{s,i}}^n, {T_{e,i}}^n, {P_i}^n]$.

In this work, we implemented a lightweight domain-specific activity chain generator utilizing the model from \cite{liao2024deep}, though the framework allows users to customize and implement alternative generation methods. This initialization process creates role-appropriate schedules that capture general behavioral patterns—students attend educational institutions, workers commute to employment locations, and retirees engage in different activity patterns. The base activity chains include essential destinations such as home locations, workplaces, and schools, establishing a realistic foundation for the simulation. These initial schedules provide a solid starting point, while subsequent modifications in response to environmental changes or simulation feedback are handled by the LLM-empowered adaptation mechanisms described in later sections.

\subsection{Simulation Tools and Integration}
After generating base activity chains for agents, MobiVerse utilizes SUMO \cite{lopez2018microscopic} as the primary traffic simulation platform to execute these plans. SUMO is an open-source, highly portable microscopic traffic simulation package designed to handle large road networks, providing comprehensive tools for modeling multimodal traffic systems. MobiVerse leverages SUMO's built-in traffic and driving behavior models to simulate each agent, enabling realistic representation of traffic dynamics at both individual and system levels. The default Krauss car-following model and LC2013 lane-changing model are adopted for driving behavior, though these models can be easily substituted with alternatives such as IDM, Wiedemann, or customized algorithms. The system handles intersection behavior through SUMO's built-in junction model, which manages gap acceptance and traffic signal interactions.

The integration with SUMO allows MobiVerse to translate activity chains generated by our hybrid model into detailed traffic simulations. We leverage SUMO's Traffic Control Interface (TraCI) API to establish bidirectional communication between our models and the traffic simulation environment. This interface enables real-time monitoring of traffic conditions and dynamic modification of agent behaviors. When environmental changes occur, such as congestion, road closures, or special events, this feedback is captured and passed to the LLM-empowered activity chain modifier module, which then adapts agent behaviors accordingly. The simulation outputs include detailed trajectory data, travel times, emissions, and other performance metrics that can be analyzed to evaluate transportation policies and infrastructure changes, creating a complete feedback loop between agent behavior and environmental conditions.

\subsection{LLM-empowered Activity Chain Modifier and Agent Adaptation}
While the lightweight domain-specific generator provide computational efficiency for base activity chain generation, they lack the flexibility to adapt to dynamic conditions during simulation. To address this limitation, MobiVerse integrates LLMs as decision-making engines for context-aware modifications to agent behavior. This integration leverages the SUMO foundation, where the traffic simulation provides the environmental context needed for adaptive decision-making.

\renewcommand{\thefigure}{3}
\begin{figure}[h]
    \centering
    \includegraphics[width=0.95\columnwidth]{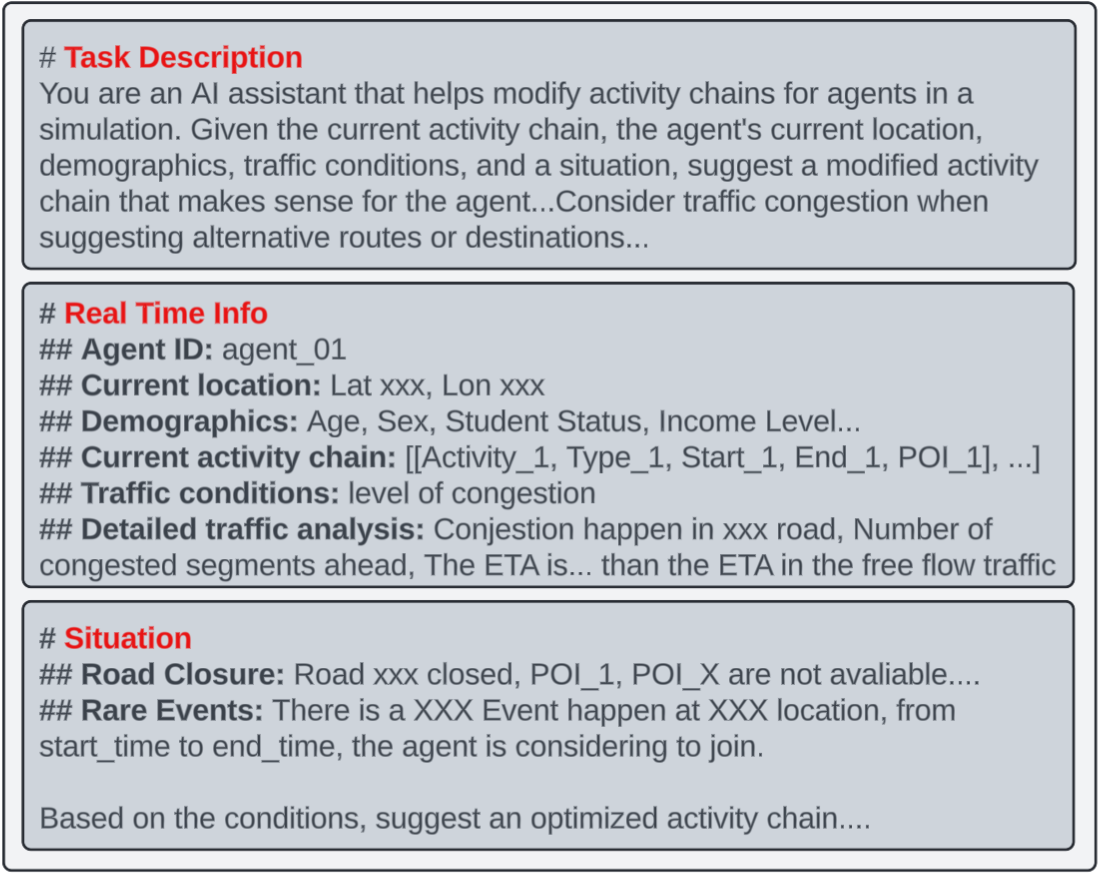}
    \caption{Example of input system prompt for LLMs.}
    \label{fig:llm_prompt}
\end{figure}

Our LLM integration follows a modular approach where the model receives comprehensive inputs about the agent's environment and characteristics. These inputs include current traffic conditions, road closures, points of interest, event information, and agent-specific information such as demographic data. This tight integration with the simulation environment enables context-aware decision making that responds to actual conditions encountered during simulation.

The LLM component is designed with flexible prompts that enable dynamic modification of agent activity schedules based on environmental feedback, as illustrated in Fig. \ref{fig:llm_prompt}. To optimize prompt management and response quality, we implement a specialized \textit{Prompt Manager} that maintains different prompt templates for various scenarios including road closures, congestion events, and special activities shown in \ref{fig:class_diagram}. Each prompt template is carefully crafted to include relevant context and constraints specific to the situation, while maintaining a consistent structure for reliable LLM responses. The prompts provide structured information about the agent's current state, including demographics, location, activity chain, and real-time traffic conditions, along with specific situations requiring adaptation.

To handle the high volume of API requests efficiently, we implement a parallel API request thread pool architecture. This system maintains a configurable pool of worker threads that process LLM API requests concurrently, with dynamic load balancing to optimize throughput. The thread pool architecture allows us to process massive modifications for thousands of agents within minutes, ensuring real-time responsiveness even in large-scale simulations. This flexibility allows us to change agents' routes in simulation environment dynamically as traffic conditions evolve.

MobiVerse also includes \textit{Handlers} to manage exceptional situations such as road closures, sporting events, or entertainment events that can significantly impact mobility patterns. For road closures, the \textit{Road Closure Handler} identifies affected agents whose destination POIs are within the closure area or whose routes include closed roads. These agents' activity chains are then modified through the LLM to find alternative destinations or routes that accommodate their needs while avoiding the affected areas. For special events like sports games or concerts, the \textit{Event handler} incorporates event-specific logic to identify potential attendees based on relevant demographic and contextual factors, enabling realistic modeling of event participation within the simulation.

\subsection{Communication and Information Flow}
\renewcommand{\thefigure}{4}
\begin{figure}[h]
    \centering
    \includegraphics[width=0.95\columnwidth]{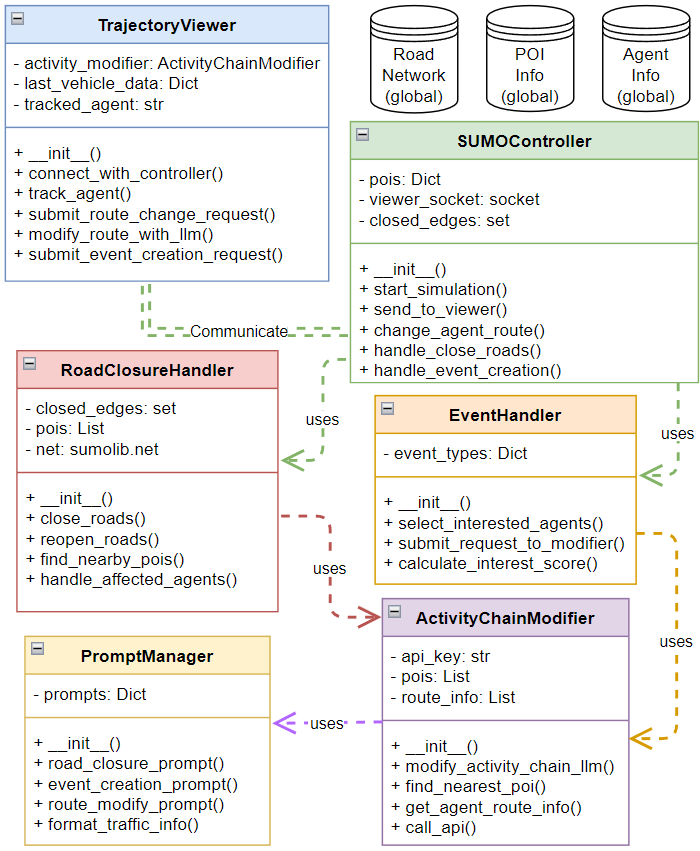}
    \caption{Class diagram of MobiVerse framework showing the main components and their relationships.}
    \label{fig:class_diagram}
\end{figure}

To facilitate the integration between the \textit{Base Activity Chain Initializer}, \textit{Simulation Environment}, and LLM-empowered \textit{Activity Chain Modifier} mechanisms described in the previous sections, MobiVerse implements a comprehensive communication system. Information flows bidirectionally through this system, creating a dynamic environment where agents can adapt to changing conditions. 

As depicted in Fig.~\ref{fig:flow_diagram_1}, the SUMO Controller executes simulation steps and collects traffic information, which is passed to the Trajectory Viewer for visualization and to event handling components for analysis. The system maintains a globally accessible agent information database that contains the latest activity chains and agent states, which is updated in real-time by all system components. At the beginning of each simulation step, the controller checks for any changes in agent activity schedules and automatically replans trajectories accordingly.

When adaptation triggers are detected (e.g., congestion thresholds, road closures, or special events), the corresponding handler identifies affected agents. The Activity Chain Modifier processes these agents at a rate of approximately 200 agents per second. This processing rate allows the simulation to continue running smoothly while modifications are being applied. The Activity Chain Modifier receives information about the affected agents and current conditions, formulating appropriate prompts for the LLM, which then generates contextually appropriate modifications to agent activity chains. These modified chains are immediately written to the global agent information database, ensuring all system components have access to the latest agent states.

The TraCI provided by SUMO serves as the primary communication channel between the traffic simulation and other components. TraCI enables retrieval of vehicle positions, speeds, and routes; modification of vehicle routes during simulation; management of traffic infrastructure (signals, road closures); and simulation control (step execution, pausing, resuming). For communication between the SUMO Controller and the Trajectory Viewer/UI components, MobiVerse implements a TCP/IP communication over sockets. This allows real-time transmission of simulation state information, user-initiated commands to modify the simulation (e.g., closing roads, creating events), and selective monitoring of specific agents or areas.

To ensure consistency across the simulation system, MobiVerse implements a synchronized time management approach where the SUMO simulation clock serves as the master time reference. All agent activities and events are scheduled relative to this reference. Simulation steps occur at configurable intervals (1 second by default), while environmental updates (congestion assessment, event processing) simultaneously with the simulation steps. The global agent information database is continuously updated to reflect the latest state of all agents, ensuring that all system components have access to consistent and up-to-date information.



\section{WESTWOOD AREA SIMULATION - A CASE STUDY}

\subsection{Experiment Setup and Data Sources}
For our experiment, we conducted a one-day simulation in the Westwood area of Los Angeles, California, involving activity planning for up to 53,000 agents, with a maximum of 20,000 agents simultaneously active in the simulation environment. The simulation area encompasses approximately 10.74 square kilometers, featuring a diverse urban landscape that includes residential neighborhoods, commercial districts, and the University of California, Los Angeles (UCLA) campus. All experiments were conducted on a personal-use level PC equipped with an Intel i7-11700F processor @ 2.50GHz, an NVIDIA RTX 3070 graphics card with 8GB VRAM, and 32GB of DDR4 memory running at 3200 MT/s. The parallel LLM API request thread pool was configured with a maximum of 100 concurrent clients to optimize throughput while staying within API rate limits.

In this study, we extract the population from the California National Household Travel Survey (NHTS) dataset~\cite{NHTS2019}. NHTS-CA provides each agent comprehensive demographic characteristics, including age, gender, employment status, income, education, etc., as agent profiles that reflect the diversity of the population in our study area. We focused on the 15 activity types aggregated in the NHTS dataset for the Los Angeles area, as shown in Table~\ref{table:activity_type_table}, which provided a comprehensive categorization of daily activities for our simulation.

\begin{table}[h]
    \centering
    \begin{tabular}{|c|c|c|c|c|c|}
        \hline
        1 & Home & 2 & Work  & 3 & School \\ \hline
        4 & Caregiving & 5 & Buy goods & 6 & Buy services \\ \hline
        7 & Buy meals & 8 & General errands & 9 & Recreational \\ \hline
        10 & Exercise  & 11 & Visit friends & 12 & Health care \\ \hline
        13 & Religious & 14 & Something else & 15 & Drop off/Pick up\\ \hline
    \end{tabular}
    \caption{Activity types aggregated in the NHTS 2017 dataset for the Los Angeles area.}
    \label{table:activity_type_table}
\end{table}

To complement the demographic data, we employed the POI dataset for the Westwood area of Los Angeles, which includes various categories of destinations such as residential areas, workplaces, educational institutions, retail establishments, restaurants, and recreational facilities. We also utilized traffic network data and POI information extracted from OpenStreetMap (OSM) \cite{haklay2008openstreetmap}, which provided comprehensive road network topology and additional location-based data. These datasets together provided the spatial framework for our simulation, allowing us to map activities to realistic locations within the urban environment and accurately represent the transportation infrastructure. Each activity in an agent's schedule is matched with a specific POI based on the agent's demographic profile, ensuring that location choices reflect realistic preferences and constraints of individuals with different characteristics.

For base activity chain generation, we implemented a deep generative model~\cite{liao2024deep}, which synthesize daily activity chain based on agents' socio-demographic profiles. This model was trained on historical mobility data and produces temporally consistent activity sequences that reflect typical human behavior patterns. The semantic-aware POI matching approach~\cite{liu2024semantic} was used for the activity-to-location matching, which leverages natural language understanding to identify suitable locations for different activities based on semantic compatibility between activity types and POI characteristics.

The simulation was initialized at 12:00 AM and ran for a full 24-hour period. Traffic conditions were updated every 5 minutes, triggering potential LLM-based decision-making for agents encountering significant changes in their environment. We implemented various scenarios to test the adaptive capabilities of our framework, including unexpected road closures, traffic congestion events, and special events at key locations within the simulation area.

\subsection{Visualization and User Interface}
MobiVerse provides an interactive visualization dashboard that enables researchers to observe and interact with the simulation. The \textit{Map View} displays the road network, vehicle positions, and points of interest, with color coding to indicate congestion levels and activity types. The \textit{Agent Inspector} allows selection of specific agents to view their demographic information, current activity, next activity, planned route, and adaptation history. For system-level analysis, the \textit{Traffic Analysis} component provides real-time statistics on network performance, including average speeds, travel times, and congestion levels across different road segments. Users can actively engage with the simulation through \textit{Intervention Tools} that enable the introduction of experimental conditions such as road closures, special events, or traffic incidents to test system responses. The visualization components communicate with the SUMO Controller through the socket interface, receiving regular updates on simulation state and transmitting user commands for simulation control or experimental intervention.

\subsection{Results}

\subsubsection{System Performance Analysis}

The initial base activity chain generation for the entire population of around 53,000 agents took 60 seconds. To evaluate computational performance, we measured runtime and speedup ratios across different agent population sizes. All measurements are based on simulating one minute of traffic (60 simulation seconds), with speedup ratio defined as simulation time divided by actual runtime. As shown in Table~\ref{table:performance_analysis}, the system maintains real-time performance with up to 20,000 active agents, achieving a speedup ratio of 1.33$\times$. For smaller populations, performance improves significantly, reaching 40$\times$ speedup with 1,000 agents.

\begin{table}[h]
    \centering
    \begin{tabular}{|c|c|c|}
        \hline
        \textbf{Active Agents} & \textbf{Runtime (s)} & \textbf{Speedup Ratio} \\ \hline
        1,000 & 1.5 & 40.00$\times$ \\ \hline
        2,000 & 2.3 & 26.08$\times$ \\ \hline
        3,000 & 7.5 & 8.00$\times$ \\ \hline
        5,000 & 13.0 & 4.61$\times$ \\ \hline
        15,000 & 30.0 & 2.00$\times$ \\ \hline
        20,000 & 45.0 & 1.33$\times$ \\ \hline
    \end{tabular}
    \caption{System performance metrics for varying agent populations.}
    \label{table:performance_analysis}
\end{table}
\vspace{-0.2cm}

The system's performance in handling dynamic events and agent replanning is detailed in Table~\ref{table:replanning_performance}. Our system processes activity sequence updates for affected agents at a rate of 2,050 agents per minute in average, followed by SUMO's route computation at around 200 agents per minute. For large-scale disruptions, the simulation will be temporarily paused to ensure all affected agents' plans are properly updated and synchronized before resuming.

\begin{table}[h]
    \centering
    \begin{tabular}{|l|c|c|}
        \hline
        \textbf{Scenario} & \textbf{SUMO Re-routing} & \textbf{LLM Replanning} \\ 
        & \textbf{(agents/min)} & \textbf{(agents/min)} \\ \hline
        Road Closure Processing & 200 & 2,100 \\ \hline
        Event Processing & 200 & 2,000 \\ \hline
    \end{tabular}
    \caption{Processing rates for LLM activity replanning and subsequent SUMO routing.}
    \label{table:replanning_performance}
\end{table}
\vspace{-0.2cm}

\subsubsection{Case Studies in Behavioral Adaptation}

The framework demonstrates sophisticated behavioral adaptation across various scenarios. In the first example shown in Fig.~\ref{fig:activity_mod}, when informed about road closure affecting the ``Whole Foods Market", the agent intelligently redirects their shopping activity to an alternative location ``Trader Joe's" while maintaining the original time slot. Similarly, upon learning that Alfred Coffe is closed due to construction, the agent adapts by selecting a new destination ``Profeta" for their food activity while preserving the same time window.

The system also demonstrates opportunistic behavior modification where an agent adjusts their evening plans to attend a soccer match at ``D Stadium" after receiving event information, showing the framework's ability to incorporate special events into existing schedules. This modification includes not only adding the new entertainment activity but also restructuring subsequent activities to accommodate the event.

When faced with significant traffic congestion (3× normal travel time to shopping destinations), the agent makes a rational time-management decision by shortening their shopping duration from 30 minutes to just 15 minutes at ``Trader Joe's", demonstrating temporal adaptation to traffic conditions while still fulfilling the essential activity.

\renewcommand{\thefigure}{5}
\begin{figure}[h]
    \centering
    \includegraphics[width=\columnwidth]{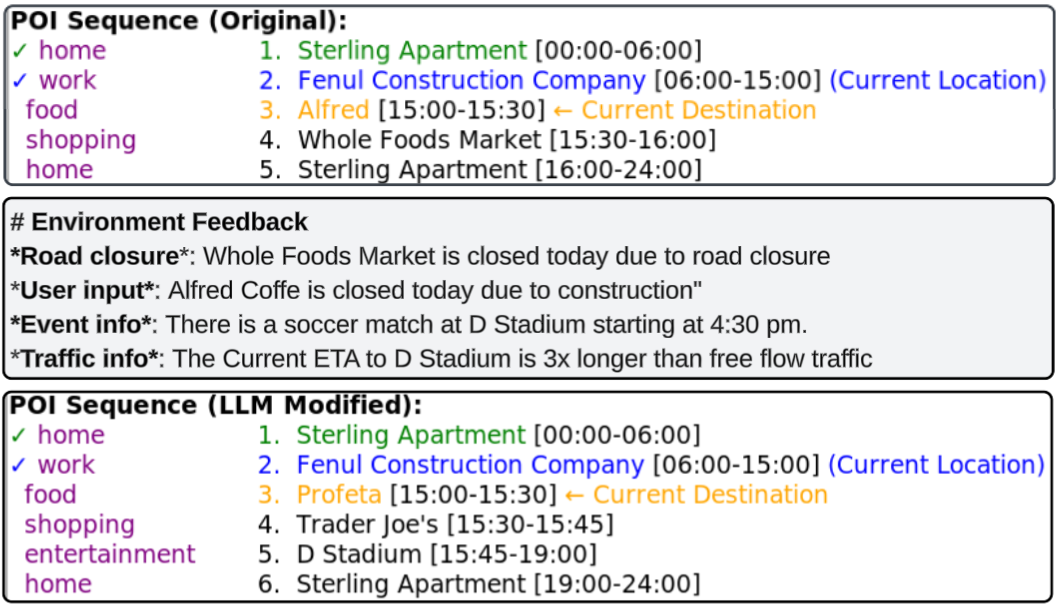}
    \caption{Examples of activity chain modifications responding to environmental triggers.}
    \label{fig:activity_mod}
\end{figure}

These examples highlight MobiVerse's capability to generate contextually appropriate responses while maintaining schedule feasibility and agent preferences. The modifications demonstrate both spatial adaptation (location changes) and temporal adjustments (duration modifications), reflecting realistic human decision-making processes when confronted with environmental constraints.

\subsubsection{Special Event Response Analysis}

To evaluate MobiVerse's capability in handling special events, we simulated the ``LA 2028 Olympic Men's Soccer Final" event at the ``Stadium" at 9:00 AM. The event was configured to accommodate 1,000 attendees, with attendee selection based on a comprehensive interest score calculation model we implemented for this case study:
\vspace{-0.2cm}
\begin{equation}
\begin{split}
\text{Interest Score} = \text{Base Interest Factor} \times \text{Age Factor} \\
\times \text{Sex Factor} \times \text{Income Factor} \times \text{Distance Factor}
\end{split}
\end{equation}

This scoring system was designed to realistically model event attendance by selecting agents with the highest interest scores up to the event's capacity. The interest score parameters were calibrated using insights from the Deloitte Football Spectator Experience Report \cite{deloitte2019football}, which provided detailed analysis of demographic factors influencing sports event attendance.

\renewcommand{\thefigure}{6}
\begin{figure}[h]
    \centering
    \includegraphics[width=0.95\columnwidth]{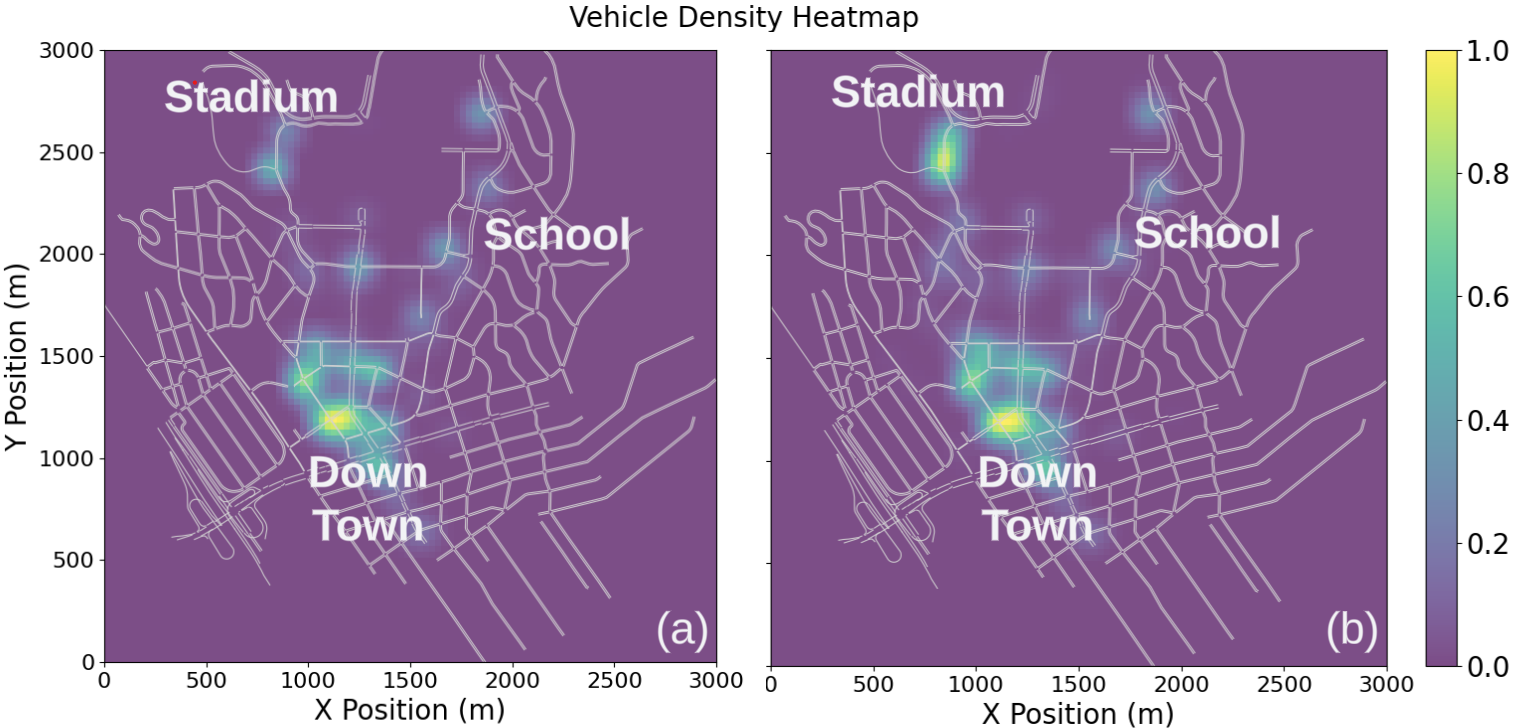}
    \caption{Vehicle density heat map at 9:30 am: (a) baseline traffic at 9:30am without event, (b) traffic at 9:30am during the Olympic soccer event}
    \label{fig:heat_map}
\end{figure}
\vspace{-0.3cm}

Fig. \ref{fig:heat_map} shows the comparison of traffic patterns with and without the event. The baseline scenario (a) shows typical morning congestion patterns in downtown and school areas, while (b) demonstrates the significant impact of the event, with notably increased traffic density around the stadium while maintaining similar patterns in other areas. The demographic distributions of selected attendees in Fig. \ref{fig:event_distribution} also closely match the survey patterns on both age and gender distributions.

\renewcommand{\thefigure}{7}
\begin{figure}[h]
    \centering
    \includegraphics[width=0.95\columnwidth]{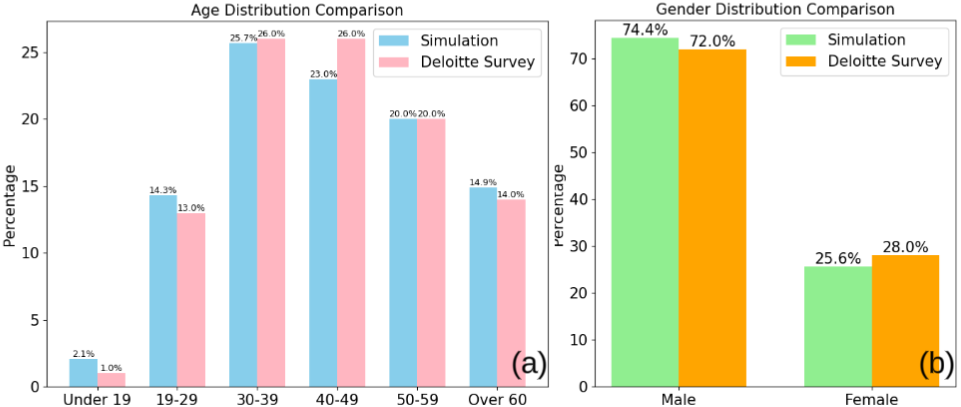}
    \caption{Event attendee demographic distribution compared with Deloitte survey: (a) age (b) gender distribution}
    \label{fig:event_distribution}
\end{figure}
\vspace{-0.5cm}

\section{CONCLUSION AND DISCUSSION}

MobiVerse integrates lightweight domain-specific generators with LLM-based decision making, enabling behavioral adaptation simulation for populations up to 53,000 agents. The hybrid approach balances computational efficiency with behavioral realism, addressing key limitations in current mobility simulation frameworks. By generating base activity chains through efficient lightweight models and applying contextual modifications via LLMs, the system provides both scalability and adaptability.

The framework offers extensive customization options, including alternative traffic simulation parameters include car-following and lane-changing models, different activity generation approaches, adjustable event handling mechanisms, and customizable LLM prompts for various adaptation scenarios. These options allow researchers to tailor the simulation to specific research questions and geographical contexts.

MobiVerse could also supports multi-day simulations for activity chains spanning beyond 24 hours while maintaining temporal consistency. Integration with SUMO provides detailed metrics including emissions and individual mobility statistics, enabling comprehensive analysis of transportation policies. The bidirectional communication between agent models and traffic simulation creates realistic feedback loops where traffic conditions influence agent decisions.

Our Westwood case study demonstrates that MobiVerse can successfully model complex urban mobility scenarios with high fidelity while maintaining reasonable computational requirements. The system effectively responds to road closures, special events, and congestion, showing potential for real-world transportation planning applications.

\section{FUTURE WORK}
Future work will focus on expanding multi-modal transportation options, enhancing computational efficiency for larger populations, and implementing more sophisticated behavioral models for specialized applications such as emergency response and sustainable transportation planning. MobiVerse represents a significant step forward in combining the efficiency of domain-specific generators with the adaptability of large language models for human mobility simulation.

\bibliographystyle{IEEEtran}
\bibliography{itsc2025_mobiverse}

\begin{thebibliography}{10}
\providecommand{\url}[1]{#1}
\csname url@samestyle\endcsname
\providecommand{\newblock}{\relax}
\providecommand{\bibinfo}[2]{#2}
\providecommand{\BIBentrySTDinterwordspacing}{\spaceskip=0pt\relax}
\providecommand{\BIBentryALTinterwordstretchfactor}{4}
\providecommand{\BIBentryALTinterwordspacing}{\spaceskip=\fontdimen2\font plus
\BIBentryALTinterwordstretchfactor\fontdimen3\font minus \fontdimen4\font\relax}
\providecommand{\BIBforeignlanguage}[2]{{%
\expandafter\ifx\csname l@#1\endcsname\relax
\typeout{** WARNING: IEEEtran.bst: No hyphenation pattern has been}%
\typeout{** loaded for the language `#1'. Using the pattern for}%
\typeout{** the default language instead.}%
\else
\language=\csname l@#1\endcsname
\fi
#2}}
\providecommand{\BIBdecl}{\relax}
\BIBdecl

\bibitem{gonzalez2008understanding}
M.~C. Gonzalez, C.~A. Hidalgo, and A.-L. Barabasi, ``Understanding individual human mobility patterns,'' \emph{nature}, vol. 453, no. 7196, pp. 779--782, 2008.

\bibitem{arentze2009need}
T.~A. Arentze and H.~J. Timmermans, ``A need-based model of multi-day, multi-person activity generation,'' \emph{Transportation Research Part B: Methodological}, vol.~43, no.~2, pp. 251--265, 2009.

\bibitem{nijland2014multi}
L.~Nijland, T.~Arentze, and H.~Timmermans, ``Multi-day activity scheduling reactions to planned activities and future events in a dynamic model of activity-travel behavior,'' \emph{Journal of Geographical Systems}, vol.~16, pp. 71--87, 2014.

\bibitem{solmaz2019survey}
G.~Solmaz and D.~Turgut, ``A survey of human mobility models,'' \emph{IEEE Access}, vol.~7, pp. 125\,711--125\,731, 2019.

\bibitem{xu2015understanding}
Y.~Xu, S.-L. Shaw, Z.~Zhao, L.~Yin, Z.~Fang, and Q.~Li, ``Understanding aggregate human mobility patterns using passive mobile phone location data: a home-based approach,'' \emph{Transportation}, vol.~42, pp. 625--646, 2015.

\bibitem{feng2020learning}
J.~Feng, Z.~Yang, F.~Xu, H.~Yu, M.~Wang, and Y.~Li, ``Learning to simulate human mobility,'' in \emph{Proceedings of the 26th ACM SIGKDD international conference on knowledge discovery \& data mining}, 2020.

\bibitem{luca2021survey}
M.~Luca, G.~Barlacchi, B.~Lepri, and L.~Pappalardo, ``A survey on deep learning for human mobility,'' \emph{ACM Computing Surveys (CSUR)}, vol.~55, no.~1, pp. 1--44, 2021.

\bibitem{liao2024deep}
X.~Liao, Q.~Jiang, B.~Y. He, Y.~Liu, C.~Kuai, and J.~Ma, ``Deep activity model: A generative approach for human mobility pattern synthesis,'' \emph{arXiv preprint arXiv:2405.17468}, 2024.

\bibitem{liu2024semantic}
Y.~Liu, C.~Kuai, X.~Liao, H.~Ma, B.~Y. He, and J.~Ma, ``Semantic trajectory data mining with llm-informed poi classification,'' in \emph{2024 IEEE 27th International Conference on Intelligent Transportation Systems (ITSC)}.\hskip 1em plus 0.5em minus 0.4em\relax IEEE, 2024, pp. 207--213.

\bibitem{gong2024mobility}
L.~Gong, Y.~Lin, Y.~Lu, X.~Han, Y.~Liu, S.~Guo, Y.~Lin, H.~Wan \emph{et~al.}, ``Mobility-llm: Learning visiting intentions and travel preference from human mobility data with large language models,'' \emph{Advances in Neural Information Processing Systems}, vol.~37, pp. 36\,185--36\,217, 2024.

\bibitem{wang2023would}
X.~Wang, M.~Fang, Z.~Zeng, and T.~Cheng, ``Where would i go next? large language models as human mobility predictors,'' \emph{arXiv preprint arXiv:2308.15197}, 2023.

\bibitem{park2023generative}
\BIBentryALTinterwordspacing
J.~S. Park, J.~C. O'Brien, C.~J. Cai, M.~R. Morris, P.~Liang, and M.~S. Bernstein, ``Generative agents: Interactive simulacra of human behavior,'' 2023. [Online]. Available: \url{https://arxiv.org/abs/2304.03442}
\BIBentrySTDinterwordspacing

\bibitem{lin2023agentsims}
J.~Lin, H.~Zhao, A.~Zhang, Y.~Wu, H.~Ping, and Q.~Chen, ``Agentsims: An open-source sandbox for large language model evaluation,'' \emph{arXiv preprint arXiv:2308.04026}, 2023.

\bibitem{jiawei2024large}
W.~JIAWEI, R.~Jiang, C.~Yang, Z.~Wu, R.~Shibasaki, N.~Koshizuka, C.~Xiao \emph{et~al.}, ``Large language models as urban residents: An llm agent framework for personal mobility generation,'' \emph{Advances in Neural Information Processing Systems}, vol.~37, 2024.

\bibitem{li2024more}
X.~Li, F.~Huang, J.~Lv, Z.~Xiao, G.~Li, and Y.~Yue, ``Be more real: Travel diary generation using llm agents and individual profiles,'' \emph{arXiv preprint arXiv:2407.18932}, 2024.

\bibitem{lopez2018microscopic}
P.~A. Lopez, M.~Behrisch, L.~Bieker-Walz, J.~Erdmann, Y.-P. Fl{\"o}tter{\"o}d, R.~Hilbrich, L.~L{\"u}cken, J.~Rummel, P.~Wagner, and E.~Wie{\ss}ner, ``Microscopic traffic simulation using sumo,'' in \emph{2018 21st international conference on intelligent transportation systems (ITSC)}.\hskip 1em plus 0.5em minus 0.4em\relax Ieee, 2018.

\bibitem{w2016multi}
K.~W~Axhausen, A.~Horni, and K.~Nagel, \emph{The multi-agent transport simulation MATSim}.\hskip 1em plus 0.5em minus 0.4em\relax Ubiquity Press, 2016.

\bibitem{he2024multi}
B.~Y. He, Q.~Jiang, J.~Ma \emph{et~al.}, ``Multi-agent multimodal transportation simulation for mega-cities: Application of los angeles,'' \emph{Procedia Computer Science}, vol. 238, pp. 736--741, 2024.

\bibitem{pinos2020automatic}
J.~Pinos, V.~Vozenilek, and O.~Pavlis, ``Automatic geodata processing methods for real-world city visualizations in cities: Skylines,'' \emph{ISPRS International Journal of Geo-Information}, vol.~9, no.~1, p.~17, 2020.

\bibitem{NHTS2019}
\BIBentryALTinterwordspacing
{National Renewable Energy Laboratory}, ``{Transportation Secure Data Center},'' Accessed Jan. 15, 2019, 2019. [Online]. Available: \url{https://www.nrel.gov/tsdc}
\BIBentrySTDinterwordspacing

\bibitem{haklay2008openstreetmap}
M.~Haklay and P.~Weber, ``Openstreetmap: User-generated street maps,'' \emph{IEEE Pervasive computing}, vol.~7, no.~4, pp. 12--18, 2008.

\bibitem{deloitte2019football}
\BIBentryALTinterwordspacing
{Deloitte Tohmatsu Consulting LLC}, ``The experience of football spectators: Creating the future of the stadium experience,'' 2019, accessed: 2025-04-21. [Online]. Available: \url{https://www2.deloitte.com/jp/en/pages/operations/articles/crm/football-spectator-experience-report.html}
\BIBentrySTDinterwordspacing

\end{thebibliography}

\end{document}